\documentclass[letterpaper, 10 pt, conference]{ieeeconf}

\IEEEoverridecommandlockouts                           

\usepackage{amssymb,amsfonts,amsmath}
\usepackage{enumerate,tikz,graphicx,mathrsfs,eucal,verbatim, bbm, derivative}
\usepackage{soul}
\usepackage{hyperref}
\usepackage{soul}
\usepackage{booktabs}
\newcommand{\ra}[1]{\renewcommand{\arraystretch}{#1}}
\usepackage{standalone}
\usepackage{algorithm}
\usepackage{algpseudocode}
\usepackage{caption}    %For images and and smaller images
\usepackage{subcaption}

\title{\LARGE \bf A Virtual Reality Framework \\ for Human-Robot Collaboration in Cloth Folding}
\author{Marco Moletta, Maciej K. Wozniak, Michael C. Welle and Danica Kragic% <-this % stops a space
\thanks{The authors are with the Robotics, Perception and Learning Lab, EECS, at KTH Royal Institute of Technology, Stockholm, Sweden
        {\tt\small moletta, maciejw, mwelle, dani@kth.se}}%
}

\begin{document}

\maketitle
\thispagestyle{empty}
\pagestyle{empty}

%%%%%%%%%%%%%%%%%%%%%%%%%%%%%%%%%%%%%%%%%%%%%%%%%%%%%%%%%%%%%%%%%%%%%%%%%%%%%%%%
\begin{abstract}
We present a virtual reality (VR) framework to automate the data collection process in cloth folding tasks. The framework uses skeleton representations to help the user define the folding plans for different classes of garments, allowing for replicating the folding on unseen items of the same class. We evaluate the framework in the context of automating garment folding tasks. A quantitative analysis is performed on three classes of garments, demonstrating that the framework reduces the need for intervention by the user. We also compare skeleton representations with RGB images in a classification task on a large dataset of clothing items, motivating the use of the proposed framework for other classes of garments.
\end{abstract}

%%%%%%%%%%%%%%%%%%%%%%%%%%%%%%%%%%%%%%%%%%%%%%%%%%%%%%%%%%%%%%%%%%%%%%%%%%%%%%%%

\section{Introduction}\label{intro}

% Problems addressed flow:
% 1. Automating industrial processes related to cloth deformable objects is necessary in the future, but for now it requires human-robot collaboration.
% 2. Folding is one example, where automation is needed but also difficult to implement due to the challenges in manipulating cloth-like objects, as discussed in literature
% 3. Also, there is the need of data to train deep learning models (which are popular in cloth manipulation lately), hence there is the need for frameworks that allow to do that. 
% 4. Virtual and augmented reality frameworks are used in industrial processes and for data collection in robotics field. But not so much for deformable object manipulation yet, hence this paper contributes on this aspect.
% 5. The aim of this work is to propose an augmented reality framework that allows to automate cloth folding. The framework allows to automatically replicate the folding plan defined by the human operator on other clothes of the same class, as well as defining new folding plans. 

There is a growing demand for the automation of garment production and recycling processes. Automating cloth manipulation tasks, such as folding or assistive dressing, could provide considerable benefits in terms of decreasing labor expenses and reducing physical effort for workers \cite{sanchez2018robotic}. Yet, robotic manipulation of deformable objects remains a significant scientific and industrial challenge \cite{zhu2022challenges}. Robots have to rely on methods that can cope with significant self-occlusions, complex interaction dynamics, motion and task planning using multimodal data \cite{tirumala2022learning}. Recent approaches resort to deep learning models to alleviate these challenges \cite{longhini2022edo, huang2022mesh}. Unfortunately, such models require a lot of training data, which is expensive to collect for deformable object manipulation. Therefore, there is a need for approaches that enable efficient data collection for garment manipulation tasks and ease collaboration between robots and humans to reduce the effort and intervention by the latter.

In this context, virtual-, augmented-, and mixed-reality (VAM) frameworks can be useful to improve human-robot collaboration, providing a common interface for the human and robot to interact \cite{wozniak2023virtual}. Such interfaces allow for more intuitive and natural communication, simplifying for the human to give instructions to the robot and for the robot to communicate its intentions \cite{chandan2021arroch}. Different frameworks have been used in robotics research for data collection and for automating and facilitating industrial processes \cite{waymouth2021demonstrating, zubizarreta2019framework}. These frameworks have shown to be preferable over traditional 2D screen interfaces, as the latter are often less intuitive and more workload-intensive for human operators \cite{hetrick2020comparing}. However, few of the current VR interfaces are tailored to clothing manipulation, which currently still requires a high level of human assistance.

\begin{figure}[!t]
    \centering
    \includegraphics[width=0.43\textwidth]{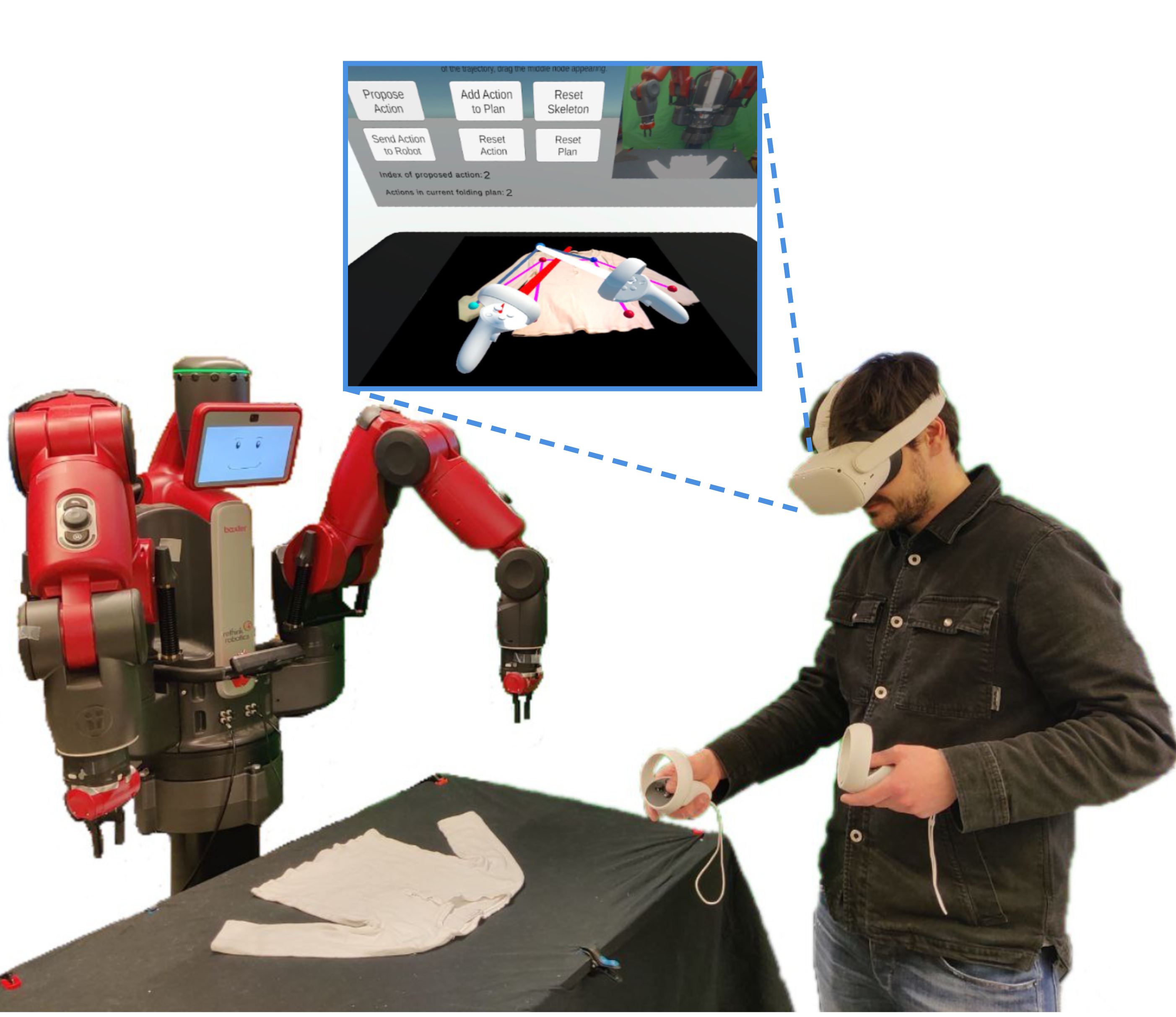}
    \caption{The proposed VR framework enables human-robot collaboration for data collection in cloth folding tasks. The user defines the folding plan by recording the sequence of pick-place actions to be executed by the robot. The plan is then replicated on unseen items of the same class.} 
    \label{fig:baxter}
    \vspace{-0.2cm}
\end{figure} 

In this paper, we present a framework for facilitating human-robot collaboration for data collection in cloth folding tasks. The concept is visualized in Fig.~\ref{fig:baxter}: the interface leverages the immersive experience provided by VR to intuitively define folding plans. The proposed framework uses skeleton representations of garments to allow for automatically replicating folding plans demonstrated by the user on unseen garments of the same class, resulting in increased automation and reduced human intervention.
% (Idea that collecting more complex/sophisticated/rich interactions it's possible to enable robots to be more human-like in their behaviours).
We evaluate the framework by defining folding plans on 3 different classes of garments, requesting the robot to replicate the plans on unseen items of the same class. We assess the level of automation and efficiency of the system by measuring the amount of intervention needed by the user. To further validate the use of skeleton representations for automating folding tasks on novel clothing items, we also employ them in an unsupervised classification task, comparing the results with RGB images.
\begin{figure*}[t]
  \centering
         \includegraphics[width=\textwidth]{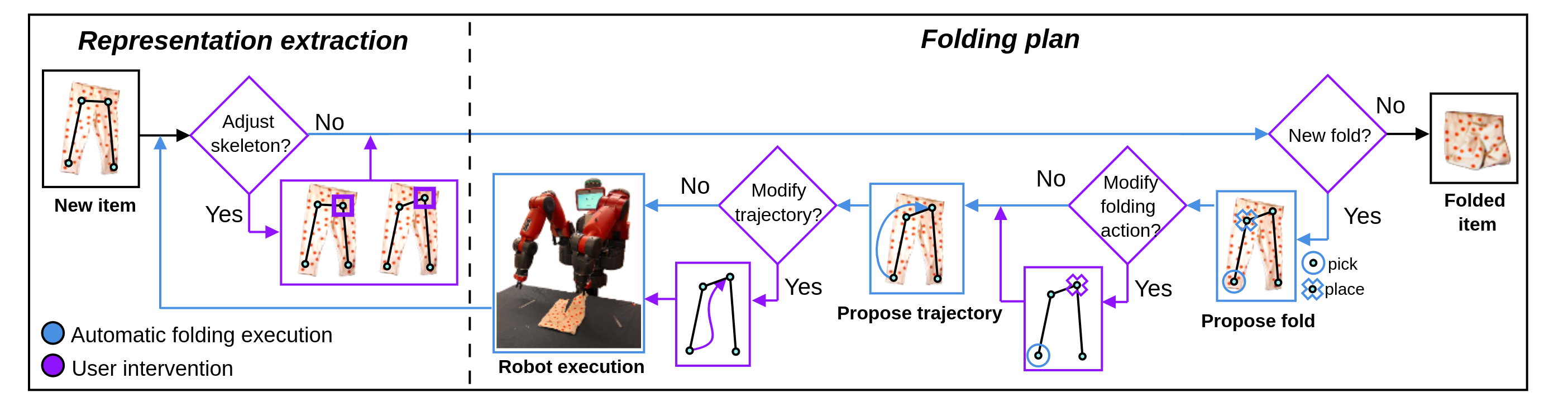}
    \caption{ The framework enables the automation of a folding task (\textit{Automatic Folding Execution}) while allowing the user to intervene (\textit{User Intervention}) if the robot proposes wrong or different folding actions than the ones desired by the user.}
    \label{fig:framework}
    \vspace{-\baselineskip}
\end{figure*}
In summary, the contributions of our work are:
\begin{itemize}
    \item A virtual reality framework based on skeleton representations that automates cloth folding for different classes of garments.
    \item A quantitative evaluation of our framework for its efficacy in automating cloth folding. \item A comparison between skeleton representations and RGB images in a garments classification task. 
\end{itemize}

\section{Related Work}
In this section, we provide an overview of the current VAM frameworks used in robotics manipulation as well as of the current approaches to cloth manipulation, highlighting the challenges that hinder the complete automation of cloth folding tasks.

\subsection{VAM frameworks in robotics}

VAM frameworks have been applied in a variety of contexts in robotics, including motion planning, control, and human-robot interaction \cite{makhataeva2020augmented, wonsick2020systematic}. Studies have also been conducted on the impact of factors such as gender on interactive robotics and teleoperation \cite{nenna2022influence}, and virtual gaming platforms have been developed to investigate interpretability and trust in human-robot collaboration \cite{mara2021cobot}.

A significant body of work has focused on the use of VR interfaces for control and teleoperation of robotic systems. In these studies, the VR interface is used to manipulate the robotic arm usually via position or velocity control \cite{barentine2021vr, xu2022shared} for example to facilitate data collection for training imitation learning policies \cite{backtomanifold}. In \cite{kennel2022multi}, a VR framework is proposed for controlling a multi-robot system composed of various manipulator robots. Additionally, VAM frameworks have been proposed to promote the collaboration between humans and robots in tasks such as handover \cite{ortenzi2022robot} and to visualize the states and intentions of the robots in delivery tasks \cite{chandan2021arroch}. These types of frameworks have also been applied in industrial settings \cite{wang2019virtual, zubizarreta2019framework} and for interactive programming of robots \cite{ostanin2018interactive}. Some of these frameworks typically also enable visualization of planning in virtual reality prior to task execution \cite{wozniak2023you, waymouth2021demonstrating}.

In the specific context of cloth manipulation, VAM frameworks are not yet well-established. Waymouth et al. \cite{waymouth2021demonstrating} did propose a 2D interface and a 3D AR framework for collecting human demonstrations for the task of folding garments. Differently from our framework, that interface neither proposed a folding plan nor allowed for replication, increasing the workload on the user. A similar framework presented in \cite{yang2016repeatable} allowed for the automation of data collection to train a deep learning architecture to perform robotics tasks, and was evaluated on folding a small piece of cloth. However, differently from our framework, that interface does not enable real-time interaction with the robot, which is necessary to enable human-robot collaboration. Borràs et al. \cite{borras2023virtual} proposed a VR framework capable of simulating realistic garments in real-time, allowing the user to collect interactions through handheld controllers. However, this framework does not integrate a physical robot and solely allows the accumulation of synthetic data. Furthermore, it lacks features to automate these interactions.

\subsection{Cloth manipulation approaches and challenges}

Manipulating clothing items poses a significant challenge due to the large number of degrees of freedom they exhibit, which causes their configuration space to be infinite-dimensional \cite{zhu2022challenges}. To overcome this challenge, current approaches to cloth manipulation resort to learning or extracting representations. For instance, some methods use particle-based representations such as graphs or meshes to discretize the cloth and approximate their dynamics \cite{lin2022learning, longhini2022edo}, while others focus on learning latent representations to reduce the dimensionality of the configuration space and enable planning \cite{lippi2022enabling, hoque2021visuospatial}.

In this context of garment manipulation, the actions are usually defined as pick-place locations on the garment, either by extracting landmarks from images \cite{gustavsson2022cloth, Schulman2013tracking, elbrechter2012folding} or by selecting nodes of the graph representations. However, since the representations are extracted from data on a particular class of garments, the learned policies and plans typically fail to generalize to unseen shapes, sizes, colors, or properties \cite{petrik2019feedback}. 
Furthermore, most folding methods assume lifting the cloth to a predefined height to complete the folding action \cite{lippi2022augment, avigal2022speedfolding} as most of the interfaces to collect folding actions do not include the possibility to vary the height or the trajectory. However, the final configuration of the garment varies considerably in relation to different heights of the pick-place actions. It is then important to include this information in the data collection process.

To enable generalization it is necessary to collect a large number of expert demonstrations for each class of garments, as well as to supervise and possibly modify the actions of the robot when deploying these models in unseen conditions. 
This motivates the framework proposed in this paper, which is explained in detail in the following section.

\section{Proposed framework}
The goal of our VR framework, shown in Fig.~\ref{fig:framework}, is to enhance the autonomy of a robot in the data collection of cloth folding task. To this end, the framework enables a human user to supervise and assist the robot during the most challenging steps currently not extensively addressed in clothing manipulation, which are the definition of a) the sequence of folding actions (folding plan), b) the pick-place locations and c) the trajectory of each folding action. 
The structure of the framework is composed of two main stages: a representation extraction stage and a folding plan stage, where the latter also enables the automatic replication of the folding task. 
These two stages are designed with the goal of reducing the workload of the user to define the folding plan for new garments of the same class. The user is still asked to approve all the steps in the pipeline to avoid errors by the robot during the folding execution. 
% The output of these modules can be used to collect representations and expert folding demonstrations.

\subsection{Representation extraction}
\begin{figure}[!ht]
    \centering
    \includegraphics[width=0.48\textwidth]{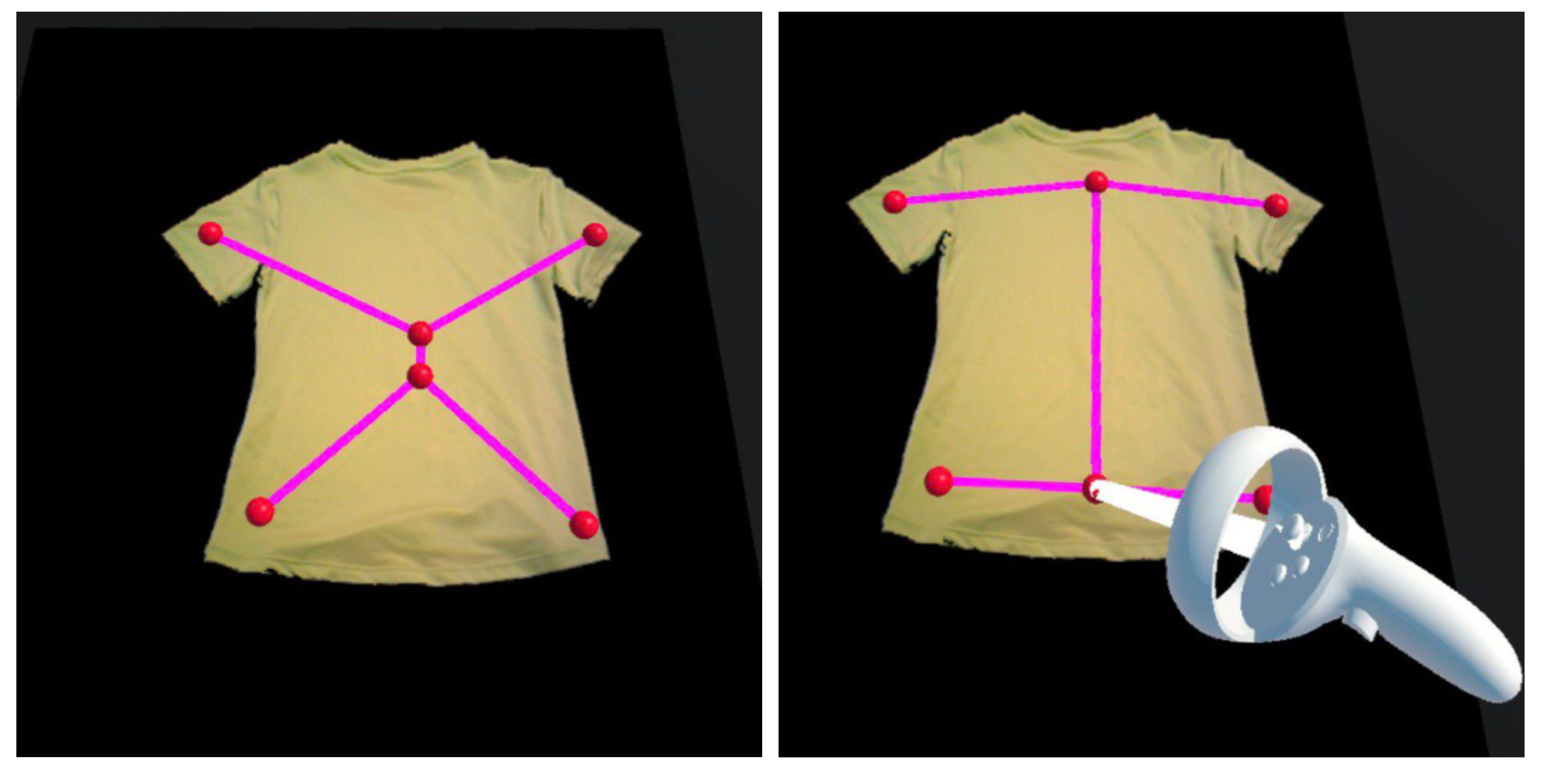}
    \caption{Modification of nodes locations through joystick interaction.} 
    \label{fig:modify_nodes}
\end{figure}

The representation extraction stage consists of creating a skeleton representation of the garment, which is a graph where nodes are used as the pick-place locations that define each folding action. The user can modify the position of the nodes in the VR environment to be in any location on the surface of the garment, as in Fig.~\ref{fig:modify_nodes}. The skeleton representation is intended as an aid to increase automation by reducing the need for the user to define all the pick-place locations of the folding actions. 

The skeleton representations are extracted from images obtained from an RGB camera. The RGB images are initially background masked and binarized. Next, we perform \textit{skeletonization} on the binary image using the thinning algorithm in \cite{Lee1994BuildingSM}, \cite{sundar2003skel} and then transform the result into a graph by creating nodes and edges as in \cite{reinders2000skeltograph}. 
The main advantage of using a skeleton representation is that the proposed action can be defined on the nodes' indices rather than pixel coordinates, making it easier to replicate it on new items given that skeletons extracted from the same class of garments most likely hold the same adjacency matrix.
% Say how indices of the nodes are given

\subsection{Folding plan} 
\begin{figure}[h]
  \centering
  \begin{subfigure}[b]{0.23\textwidth}
    \centering
    \includegraphics[width=\textwidth]{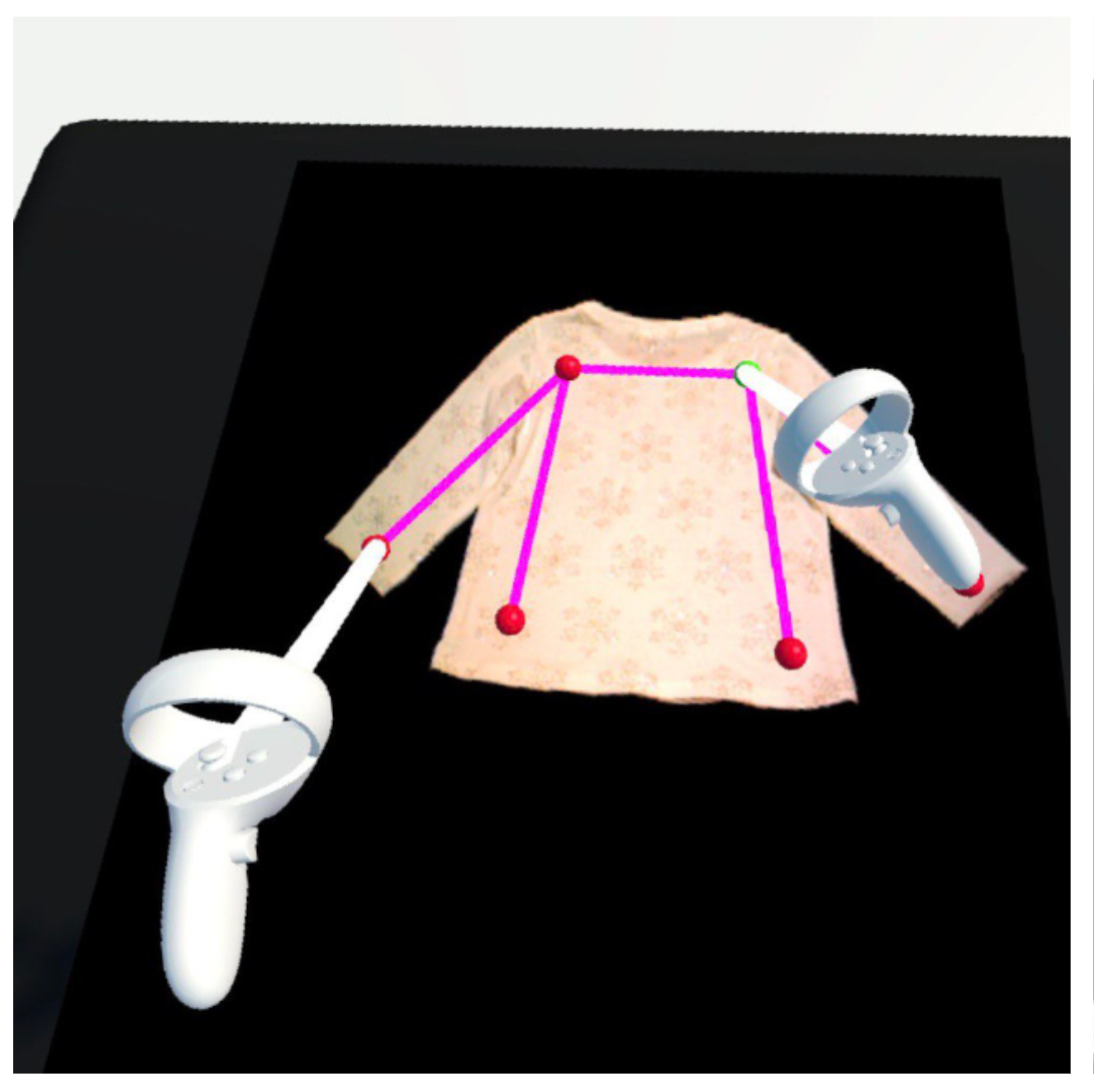}
    \caption{}
    \label{fig:traj_subfig_a} % Optional. Only if you want to refer to this subfigure elsewhere.
  \end{subfigure}
  \hfill
  \begin{subfigure}[b]{0.23\textwidth}
    \centering
    \includegraphics[width=\textwidth]{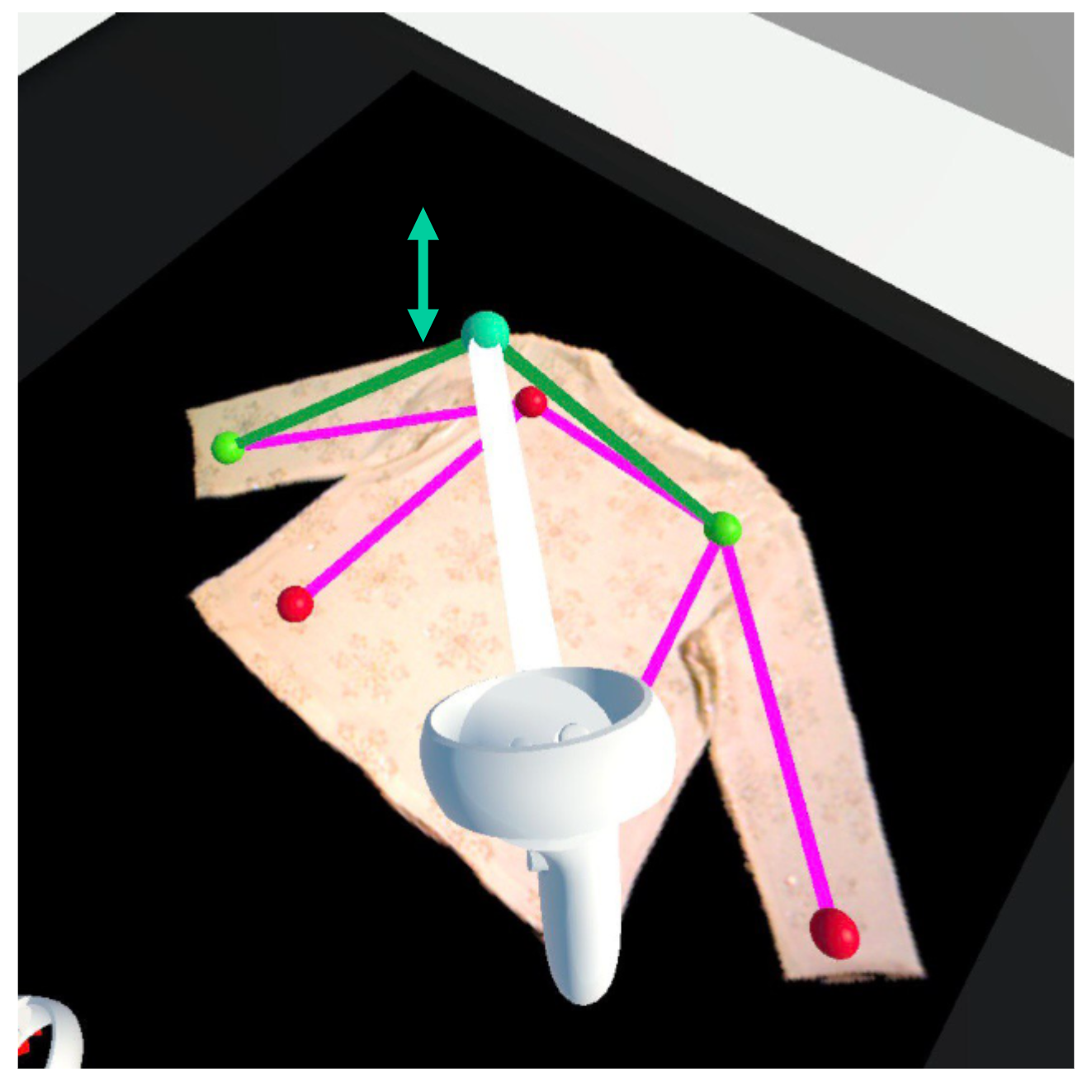}
    \caption{}
    \label{fig:traj_subfig_b} % Optional. Only if you want to refer to this subfigure elsewhere.
  \end{subfigure}
  \caption{(a) The folding action is defined by selecting the pick-place nodes. (b) The trajectory can be modified by increasing or decreasing the height of the middle-node.}
  \label{fig:traj} % Optional. Only if you want to refer to this whole figure elsewhere.
\end{figure}

\begin{figure}[!ht]
    \centering
    \includegraphics[width=0.48\textwidth]{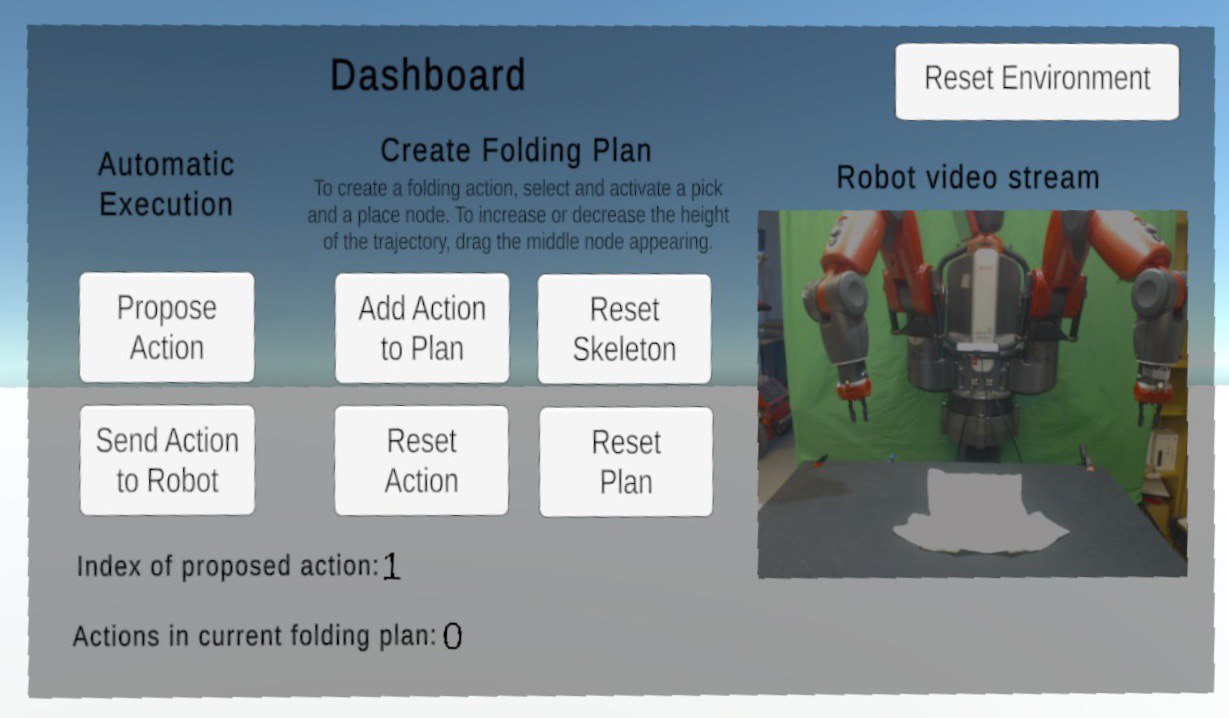}
    \caption{The dashboard the user uses to create, reset, and save folding actions and plans. The interface allows to visualize the number of actions in the saved folding plan, as well as the real-time video streaming of the robot. } 
    \label{fig:interface}
\end{figure}

\begin{figure}[!ht]
    \centering
    \includegraphics[width=0.3\textwidth]{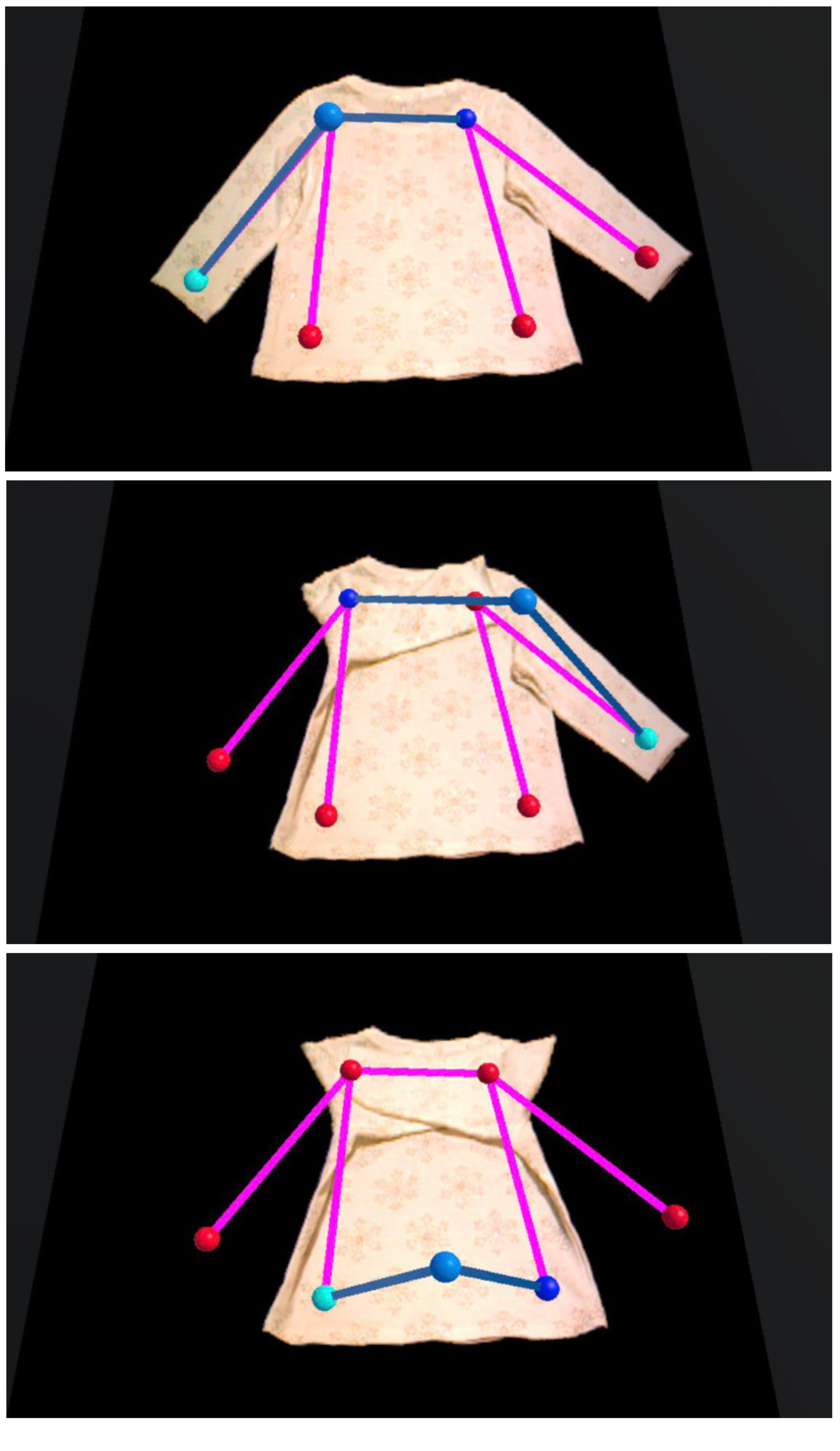}
    \caption{Different folding actions proposed by the robot during the \textit{automatic folding execution}, pick-nodes displayed in light blue, place-nodes in blue.} 
    \label{fig:proposed_action}
\end{figure}

A folding plan is a sequence of consecutive pick-place folding actions defined by the human and executed by the robot. A folding plan is defined for a specific class of garments. To create a new folding action, the user selects and activates a pick- and a place-node of the skeleton (Fig.~\ref{fig:traj_subfig_a}). This triggers the spawning of a middle-node in between the two, which allows the user to increase or decrease the height of the trajectory of the end effector of the robot (Fig.~\ref{fig:traj_subfig_b}). Once the action is defined, it can be saved in the current folding plan by using the \textit{Add Action to Plan} button of the interface, visible in Fig.~\ref{fig:interface}.
The folding plan can then be replicated on new garments of the same class, which is called \textit{Automatic folding execution}. In the replication process, the framework suggests a series of pick-place actions and relative trajectories based on the new representation extracted from the unseen item to be folded  (Fig.~\ref{fig:proposed_action}). The user can ask the robot to propose a folding action using the \textit{Propose Action} button. The action can be approved by pressing \textit{Send Action to Robot} or modified by the user before execution. The modifications consist of both resetting and changing the pick-place nodes of the action (\textit{Reset Action}) as well as adjusting the intermediate position of the end-effector throughout the movement (the "height" of the trajectory).

% \begin{figure}[!ht]
%     \centering
%     \includegraphics[width=0.5\textwidth]{figures/traj_sequence.png}
%     \caption{Left: the folding action is defined by selecting the pick-place nodes. Right: the trajectory can be modified by increasing or decreasing the height of the middle-node.} 
%     \label{fig:traj}
% \end{figure} 
% Explain on why not imitation learning as skeleton are easier to extract from the image.

\subsection{Hardware setup}
We implemented the user interface in Unity 2020.3 and deploy it on an Oculus Quest 2 headset. We use a Baxter robot, which was already previously used in cloth folding manipulation tasks \cite{lippi2022enabling}. A ROS (Robot Operating System) node is used to exchange information between the robot and the user through TCP (Transmission Control Protocol).

\section{Experiments}

In this section, we evaluate how the proposed framework can facilitate the automation of folding different classes of garments. In particular:  
\begin{enumerate}
    \item We provide quantitative results on the success of the robot and the amount of intervention needed by the user when folding plans defined for a cloth of a specific class are automatically replicated on other clothes of the same class.
    \item To support the hypothesis that skeleton representations are different enough across different classes to allow folding plans to be replicated, we provide the results of an unsupervised classification task on a dataset of clothing items and compare them to RGB images.
\end{enumerate}

\subsection{Automatic folding execution and User intervention}

% {\red since saving plan, why not undo and try to unfold following the plan back?}

%save one plan on one item - replicate plan using interface on same item and new items for 5 times. 

% short sleeve top:
% saving 2 plans on green 3 steps
% replicate on same with slightly different pose

% proposal considered correct if all actions are proposed correctly, meaning if the representation is extracted correctly.

\begin{figure}[!t]
    \centering
    \includegraphics[width=0.5\textwidth]{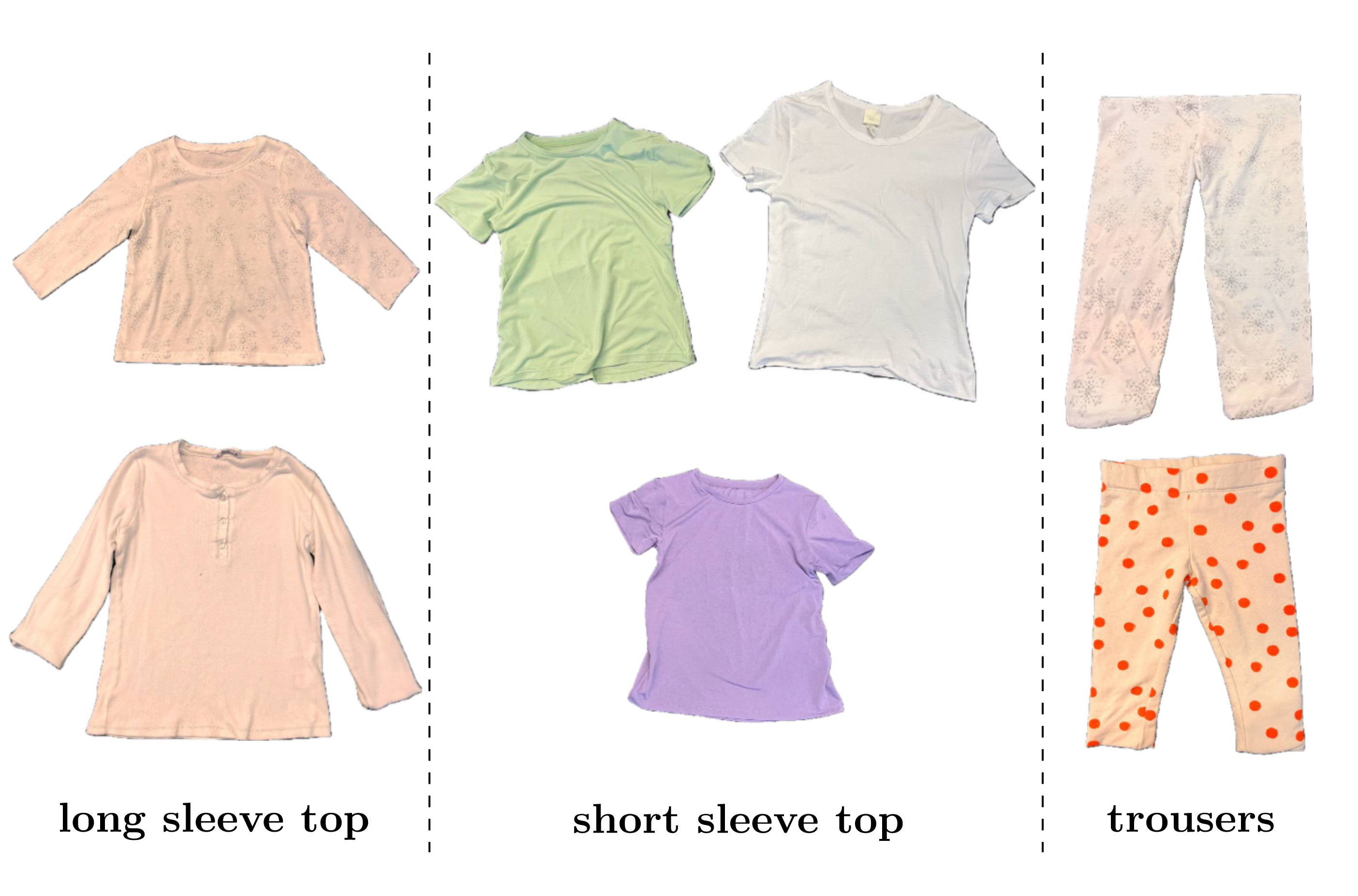}
    \caption{The objects set used in the automation experiment:  $2$ \textit{long sleeve top} (small, large), $3$ \textit{short sleeve top} (green, white, purple) and $2$ \textit{trousers} (white, pois).} 
    \label{fig:objects_set}
\end{figure}

The objective of the experiment is to assess to which extent interventions by the user are needed when using the framework in \textit{automatic folding execution}. We evaluate the automation capability by defining and replicating different folding plans on 3 different classes of clothing items, namely \textit{short sleeve top}, \textit{long sleeve top} and \textit{trousers}, following the taxonomy of Deepfashion2 \cite{ge2019deepfashion2}. The object set used in the experiments is visible in Fig.~\ref{fig:objects_set}. 

\begin{algorithm}[!t]
\caption{Evaluation procedure of automatic execution.}\label{alg:autom}
\label{tab:auto_results}
\begin{algorithmic}[1]
    \State \textit{R:} Robot, \textit{U:} User.
    \For{every Class of Garments}
        \State \textit{U:} defines exemplary Folding Plan
            \For {every Garment in Class} 
                \State Garment is placed in front of Robot.
                \State \textit{R:} displays Skeleton Representation. 
                \If{\textit{U:} accepts Skeleton Representation}
                    \For {every Action in Folding Plan}
                        \State \textit{U:} Adjusts nodes position (if needed).
                        \State \textit{R:} proposes Folding Action.
                        \If{\textit{U:} accepts Folding Action}
                            \State \textit{R:} performs Folding Action. 
                        \Else
                            \State \textbf{Proposal error}.
                            \State \textit{U:} defines Folding Action.
                            \State \textit{R:} performs Folding Action.
                        \EndIf
                    \EndFor
                \Else
                    \State \textbf{Representation error}.
                    \State \textit{U:} adjusts Representation.
                    \State \textit{U:} defines new Folding Plan.
                    \State \textit{R:} executes Folding Plan.
                \EndIf
            \EndFor  
        \EndFor
\end{algorithmic} 
\end{algorithm}

The evaluation procedure is visible in Algorithm~\ref{alg:autom}. The procedure is as follows: for each new class of garments, an exemplary folding plan is created and saved by the user on one of the items, which is then replicated on all the items in the class. When a new item is placed in front of the robot to be folded, the skeleton representation is visualized in the VR interface and the user asks the robot to propose the actions present in the folding plan. For each of the actions in the folding plan, if the action displayed is the same that the user defined in the ideal folding plan, the user will accept it and robot executes it, otherwise the user will intervene and modify it. 

The accuracy results are reported in Table~\ref{tab:auto_results}. For each item in each class of garments, the folding plan is replicated $3$ times, resulting in $3\cdot|\text{folding plan}|$ proposals of folding actions by the robot. The errors can be due to both failures in the representations extraction and in the proposal of the folding actions to be executed, where an error means a user intervention. When a failure happens in the representation extraction stage, the user has to intervene and redefine the whole folding plan, as the folding action proposals will not match the ones defined in the exemplary folding plan. Failures due to the robot execution (dropping or not releasing the garment, inverse kinematics errors, etc.) are not reported in the results. 

From the results we can conclude that the user intervention is limited, meaning that the framework enables a considerable increase in the level of automation in cloth folding tasks. A video showing different automatic folding executions is present in the supplementary material.

\begin{table}
\centering
\ra{1.3}
\caption {Results of automatic folding execution.}
\label{tab:manipulation_results1}

\begin{tabular}{lllll} %@{}c@{}c@{}c@{}c@{}
\toprule
  \textbf{Class} & \textbf{\text{Folding}} & \textbf{Item} & \textbf{Representation} & \textbf{Proposal}   \\ 
                 &        \textbf{\text{Plan}}                          &               & \textbf{Accuracy}         & \textbf{Accuracy}   \\       
        
\midrule
% long sleeve top     &  4 steps - 80\%                                        \\ \hline
short sleeve top & 2  &  purple & 3/3     & 6/6         \\ 
                 &          &  green  & 3/3   & 6/6            \\ 
                 &          &  white  & 2/3   & 4/6            \\
\midrule
short sleeve top & 4  &  purple  & 3/3     & 12/12         \\ 
                 &           &  green & 3/3    & 11/12           \\ 
                 &           &  white  & 2/3    & 8/12           \\
\midrule
long sleeve top & 3   &  large & 2/3       & 6/9        \\ 
                &            &  small & 3/3   & 8/9            \\ 
\midrule
trousers        & 2      &  pois &   1/3    & 2/6      \\ 
                &            &  white &   3/3    & 6/6      \\
\bottomrule
\end{tabular}
\end{table}

\subsection{Garments unsupervised classification task}

% To validate further that a folding plan defined on a garment is replicable on unseen items of the same class, we want to estimate to which extent the skeleton representations on which the folding plan is defined are different among different classes of garments not present in the object set in Fig.~\ref{fig:objects_set}. To do that, we evaluate the skeleton representations in an unsupervised classification task, comparing the results with original RGB images. 
Due to the vast diversity and complexity of garment classification, it would be unrealistic to test all the possible classes on our framework, especially considering the lack of uniform labeling standards for clothing items. Our aim is to determine the potential replicability of a defined folding plan for unseen clothing items within the same class. This involves measuring differences in skeleton representations across diverse garment classes not depicted in Fig.~\ref{fig:objects_set}. High accuracy in distinguishing item classes solely based on skeleton representations increases the likelihood of replicating the plan on novel class items. To make our analysis not biased on specific labels, we examine skeleton representations in an unsupervised classification task, comparing results with original RGB images. We resort to two contrastive learning frameworks: SimCLR \cite{chen2020simple}, and MVGRL \cite{MVGRL}, respectively, for images and skeleton representations.  

The dataset used for the evaluation is a subset of the Deepfashion2 dataset \cite{ge2019deepfashion2}.
The Deepfashion2 dataset contains 491k images of $13$  categories of clothing items, which include attributes such as \textit{scale, occlusion, zoom-in, viewpoint}.
For our evaluation, we removed four under-represented categories (\textit{short sleeve outwear}, \textit{sling}, \textit{long sleeve dress} and \textit{sling dress}) and used images with attributes: \textit{scale = moderate, occlusion = no/slight, zoom-in = no} and \textit{viewpoint = frontal}. The images in the dataset are downsampled and padded to 160*160 pixels. In total, the final dataset consists of $25102$ images divided into $9$ classes, where $85\%$ are used for training and $15\%$ for testing.  The composition of the dataset can be seen in Table~\ref{tab:categories}.

\begin{table}[!t]
\centering
\ra{1.3}
\caption{classes and composition of the training and test sets (number of samples - percentage).}
\label{tab:categories}
\begin{tabular}{ @{}llll@{}}
\toprule
 \textbf{Id} & \textbf{Name} & \textbf{Training set} & \textbf{Test set} \\
 \midrule
 0 & short sleeve top & 3999 - 18.4\% & 661 - 19.8\% \\
 1 & long sleeve top & 3999 - 18.4\% & 661 - 19.8\%\\
 2 & long sleeve outwear & 1877 - 8.6\% & 257 - 7.7\%\\
 3 & vest & 1645 - 7.5\% & 226 - 6.7\%\\
 4 & shorts & 1527 - 7.0\% & 127 - 3.8\%\\
 5 & trousers & 3059 - 14.0\% & 176 - 5.2\%\\
 6 & skirt & 983 - 4.5\% & 282 - 8.4\%\\
 7 & short sleeve dress & 2618 - 12.0\% & 567 - 16.9\%\\
 8 & vest dress & 2070 - 9.5\% & 380 - 11.4\%\\
\bottomrule
\end{tabular}
\end{table}

\begin{table}[!tb]
\centering
\ra{1.3}
\caption{classification results using the KNN evaluation protocol for SimCLR, and the linear evaluation protocol for MVGRL.}
\label{tab:classification_results}
\begin{tabular}{ @{}lcc@{}}
\toprule
\textbf{Model}                & \textbf{Top 1 Acc.} & \textbf{Top 5 Acc.} \\ 
\midrule
Original-RGB (SimCLR)      & 54.1 \%         &    84.8 \%      \\ 
%Binary (SimCLR) & 65.0 \%        &    85.2 \%      \\ 
Skeleton (MVGRL)       & 52.4 \%          &    90.7 \%      \\ 
Random                  & 10.9 \%   & 60.2 \%  \\
\bottomrule
\end{tabular}
\end{table}

Since in contrastive learning the composition of data augmentations plays a critical role in increasing the performance of the extracted representation, we provide details on the types of augmentations used both for images and skeleton representations. 
From \cite{chen2020simple}, for original-RGB we apply \textit{random-resize-crop}, \textit{color-jitter}, \textit{Random horizontal flip} and \textit{random grayscale} augmentations. 
From \cite{MVGRL}, we apply \textit{diffusion}, which augments the adjacency matrix of the skeleton graph with additional edges. We also employ \textit{Horizontal Flip} and \textit{Vertical Flip}, which mirror the graph respectively on the horizontal and vertical axis. We experimented also with adding noise but found that it decreases performance, which is also pointed out in \cite{MVGRL}.

% The aforementioned augmentations are the ones used in the performance comparison between representations obtained with SimCLR (original RGB) and representations obtained with MVGRL (skeletons), reported in Table~\ref{tab:classification_results}.
To make the comparison as fair as possible, we present and evaluate the results from the best-performing augmentations and the best-performing evaluation protocols from both types of representations, which in our case is the KNN evaluation protocol for SimCLR \cite{chen2020simple} and the linear evaluation protocol for MVGRL \cite{MVGRL}.
In this experiment, the batch size is set to $64$, and both models are trained for $1000$ epochs. Both models are trained with the loss functions used in the original papers (NT-Xent loss \cite{chen2020simple} for SimCLR and Jensen-Shannon divergence (JSD) for MVGRL \cite{MVGRL}.
For comparison, a random classifier is also implemented as a baseline model, which randomly assigns class labels to instances in the test set without any training. We report the Top 1 and Top 5 accuracies of all the models in Table~\ref{tab:classification_results}.
% For comparison, we also report the Top 1 and Top 5 accuracies of random classification performed on the test dataset.
% We calculate the top-1 accuracy, which represents the proportion of correct predictions among all predictions, and the top-5 accuracy, which considers whether the true label is within the top 5 predicted labels. These accuracy metrics are calculated by comparing the predicted labels to the true labels.

% The best results are achieved by SimCLR on binary images. 
The classification results for skeleton representations are comparable to original RGB images and considerably better than random. This suggests that skeleton representations of different garments are as different as original RGB images, suggesting that folding plans are likely to be reproducible also on other garments not evaluated in the automatic folding experiment. Moreover, these insights open the possibility of integrating a classification framework based on images or skeletons in future work, for example, to propose some default folding plan based on the class of garment. 
% (dataset with instances of other different  classes of garments for further use of the framework, also for suggesting that this representations could be used collected together with the interactions)}.

\section{Limitations and Future work}

One current limitation of the framework is that if the extracted skeleton differs for items of the same class, the folding plan is not automatically replicable and require some user intervention. One way to address this may be to allow rather simple skeletons for each class by limiting the number of skeleton nodes. We also plan to run a user study to collect feedback about the interface and test the automation capabilities of the framework more thoroughly with novice users. 
We also plan to look into defining bimanual folding actions by pick-placing $2$ nodes simultaneously, to further improve the data generation aspect by allowing more sophisticated folding trajectories.

\section{Conclusion}
We presented a VR framework to automate the data collection process of cloth folding tasks. The framework relies on skeleton representations to help the user to define the folding plan for different classes of garments, allowing the replication of the folding plan on unseen items of the same class.
We quantitatively evaluate our framework for its efficacy in automating cloth folding and we compare skeleton representations with RGB images in a garments unsupervised classification tasks. We conclude that the framework enhances the automation of cloth folding tasks, and the classification results suggest that this could be valid also for classes of garments not evaluated in the automatic folding experiments.
We also plan to further expand the framework for the automation of flattening tasks, with the aim of creating a complete interface for the collection of cloth manipulation data in a more automated manner through human-robot collaboration.

% Also planning on saving how the representation is modified by the user (how nodes are moved), for example training a network that learns from the user their preferences on how to fold.

% \bibliographystyle{IEEEtranS.bst}
\bibliographystyle{ieeetr}
\bibliography{references}

\end{document}